\documentclass[10pt,twocolumn,letterpaper]{article}

\usepackage{cvpr}              

\usepackage{graphicx}
\usepackage{amsmath}
\usepackage{amssymb}
\usepackage{booktabs}
\usepackage{comment}
\usepackage{tabularx}
\usepackage{multirow}
\usepackage{makecell}
\usepackage{marvosym}
\usepackage{capt-of}
\usepackage[accsupp]{axessibility}
\usepackage{multicol}
\usepackage{blindtext}
\usepackage{xcolor}

\newcolumntype{Y}{>{\centering\arraybackslash}X}

\usepackage[pagebackref,breaklinks,colorlinks]{hyperref}

\usepackage[capitalize]{cleveref}
\crefname{section}{Sec.}{Secs.}
\Crefname{section}{Section}{Sections}
\Crefname{table}{Table}{Tables}
\crefname{table}{Tab.}{Tabs.}

\def\cvprPaperID{4299} 
\def\confName{CVPR}
\def\confYear{2023}

\begin{document}

\title{Im2Hands: Learning Attentive Implicit Representation \\of Interacting Two-Hand Shapes}

\author{Jihyun Lee$^1$
\qquad
Minhyuk Sung$^1$
\qquad
Honggyu Choi$^1$
\qquad
Tae-Kyun Kim$^{1,2}$\\
$^1$ KAIST \qquad $^2$ Imperial College London}

\twocolumn[{%
\renewcommand\twocolumn[1][]{#1}%
\maketitle
\begin{center}
\centering
\captionsetup{type=figure}
\includegraphics[width=\textwidth]{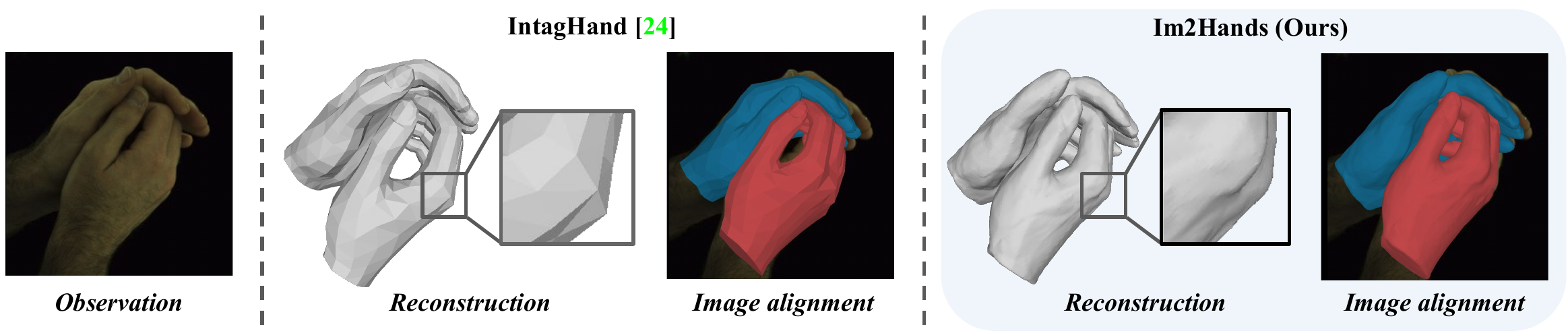}
\caption{\textbf{Reconstructed two-hand shapes from an RGB image.} We present Im2Hands, the first neural implicit representation for two interacting hands. Compared to the existing mesh-based two-hand reconstruction method (IntagHand~\cite{li2022interacting}), Im2Hands effectively captures the fine-grained geometry of two hands with higher shape-to-image coherency. The above results were produced from a single RGB image input, where ours utilized an off-the-shelf two-hand keypoint estimation method (DIGIT~\cite{fan2021learning}) to condition the articulated occupancy.}
\label{fig:teaser_image}
\end{center}
}]

\begin{abstract}
\vspace{-0.6\baselineskip}

   We present \textbf{Im}plicit \textbf{Two} \textbf{Hands} (Im2Hands), the first neural implicit representation of two interacting hands. Unlike existing methods on two-hand reconstruction that rely on a parametric hand model and/or low-resolution meshes, Im2Hands can produce fine-grained geometry of two hands with high hand-to-hand and hand-to-image coherency. To handle the shape complexity and interaction context between two hands, Im2Hands models the occupancy volume of two hands -- conditioned on an RGB image and coarse 3D keypoints -- by two novel attention-based modules responsible for (1) initial occupancy estimation and (2) context-aware occupancy refinement, respectively. Im2Hands first learns per-hand neural articulated occupancy in the canonical space designed for each hand using query-image attention. It then refines the initial two-hand occupancy in the posed space to enhance the coherency between the two hand shapes using query-anchor attention. In addition, we introduce an optional keypoint refinement module to enable robust two-hand shape estimation from \emph{predicted} hand keypoints in a single-image reconstruction scenario. We experimentally demonstrate the effectiveness of Im2Hands on two-hand reconstruction in comparison to related methods, where ours achieves state-of-the-art results. Our code is publicly available at \url{https://github.com/jyunlee/Im2Hands}.
   
\vspace{-2\baselineskip}
\end{abstract}

\vspace{\baselineskip}
\section{Introduction}
\label{sec:intro}

Humans use hand-to-hand interaction in everyday activities, which makes modeling 3D shapes of two interacting hands important for various applications (e.g., human-computer interaction, robotics, and augmented or virtual reality). However, the domain of two-hand shape reconstruction remains relatively under-explored, while many existing studies have put efforts into \emph{single-hand} reconstruction from RGB~\cite{ge20193d, kulon2020weakly, zhou2020monocular, lin2021end, baek2019pushing, baek2020weakly, boukhayma20193d}, depth~\cite{wan2020dual}, or sparse keypoints~\cite{karunratanakul2021skeleton, zhou2020monocular}. These single hand-based methods are not effective when directly applied for interacting two-hand reconstruction, since it introduces additional challenges including inter-hand collisions and mutual occlusions. 


Recently, few learning-based methods on two-hand shape reconstruction~\cite{li2022interacting, zhang2021interacting, rong2021monocular} have been proposed since the release of the large-scale interacting hand dataset (i.e., InterHand2.6M~\cite{moon2020interhand2}). Two-Hand-Shape-Pose~\cite{zhang2021interacting} and IHMR~\cite{rong2021monocular} reconstruct two-hands by estimating MANO~\cite{romero2017embodied} parameters, which are later mapped to triangular hand meshes using a pre-defined statistical model (i.e., MANO). IntagHand~\cite{li2022interacting} directly regresses a fixed number of mesh vertex coordinates using a graph convolutional network (GCN). These methods mainly model the shape of two interacting hands based on a low-resolution mesh representation with a fixed topology of MANO (please refer to Figure~\ref{fig:teaser_image}). 


In this paper, we present \textbf{Im}plicit \textbf{Two} \textbf{Hands} (Im2Hands), the first neural implicit representation of two interacting hands. Unlike the existing mesh-based two-hand reconstruction methods, Im2Hands can capture the fine-grained geometry of two interacting hands by learning a \emph{continuous} 3D occupancy field. Im2Hands (1) produces two-hand meshes with an arbitrary resolution, (2) does not require dense vertex correspondences or statistical model parameter annotations for training, and (3) learns output shapes with precise hand-to-hand and hand-to-image alignment. As two interacting hands are highly articulated objects, we take inspiration from recent neural \emph{articulated} implicit functions~\cite{deng2020nasa, karunratanakul2021skeleton, corona2022lisa, noguchi2021neural, saito2021scanimate, mihajlovic2022coap} that learn an implicit geometry leveraging the object canonical space computed from an input pose observation. Our two-hand articulated implicit function is also driven by input pose and shape observations, which are represented as sparse 3D keypoints and an RGB image, respectively. 

To effectively handle the shape complexity and interaction context between two hands, Im2Hands consists of two novel attention-based modules responsible for initial hand occupancy estimation and context-aware two-hand occupancy refinement, respectively. The initial occupancy estimation network first predicts the articulated occupancy volume of each hand in the canonical space. Given a 3D query point, it (1) performs query canonicalization using the keypoint encoder of HALO~\cite{karunratanakul2021skeleton} to effectively capture \emph{pose-dependent hand deformations} and (2) extracts a hand shape feature using our novel query-image attention module to model \emph{shape-dependent hand deformations}. As it is non-trivial to model two-hand interaction while learning in the canonical space defined for each hand, our context-aware occupancy refinement network modifies the initial two-hand occupancy in the original posed space to enhance hand-to-hand coherency. Given the initial two-hand shape represented as anchored point clouds, it uses query-anchor attention to learn a refined two-hand occupancy in a context-aware manner. Furthermore, we consider a practical scenario of two-hand reconstruction using Im2Hands from single images, where no ground truth keypoints are observed as inputs to our method. To this end, we introduce an \emph{optional} input keypoint refinement network to enable more robust two-hand shape reconstruction by alleviating errors in the input 3D keypoints predicted from an off-the-shelf image-based two-hand keypoint estimation method (e.g., \cite{fan2021learning, moon2020interhand2, kim2021end, li2022interacting, zhang2021interacting}).

Overall, our main contributions are summarized as follows:

\begin{itemize}

    \item We introduce Im2Hands, the first neural implicit representation of two interacting hands. Im2Hands reconstructs resolution-independent geometry of two-hands with high hand-to-hand and hand-to-image coherency.
    
    \item To effectively learn an occupancy field of the complex two-hand geometries, we propose two novel attention-based modules that perform (1) initial occupancy estimation in the canonical space and (2) context-aware occupancy refinement in the original posed space, respectively. We additionally introduce an \emph{optional} keypoint refinement module to enable more robust two-hand shape estimation using a single image input.
    
    \item We demonstrate the effectiveness of Im2Hands in comparison to the existing (1) two-hand mesh-based and (2) single-hand implicit function-based reconstruction methods, where Im2Hands achieves state-of-the-art results in interacting two-hand reconstruction.
    
\end{itemize}

\section{Related Work}
\label{sec:related_work}

\noindent \textbf{Single-Hand Reconstruction.}
Methods for single-hand reconstruction have been actively investigated in the past decades. Most existing deep learning-based approaches either reconstruct hand poses represented as 3D keypoints~\cite{ge2016robust, iqbal2018hand, moon2018v2v, zimmermann2017learning, cai2018weakly, simon2017hand, spurr2020weakly, wu2005analyzing}, estimate MANO parameters~\cite{romero2017embodied, baek2019pushing, baek2020weakly}, or directly regress mesh vertex coordinates~\cite{lin2021end, ge20193d, kulon2020weakly, wan2020dual}. Inspired by the recent success of implicit representations in modeling human bodies~\cite{deng2020nasa, noguchi2021neural, saito2021scanimate, mihajlovic2022coap}, few recent works~\cite{karunratanakul2021skeleton, corona2022lisa} also employ neural implicit functions for single-hand reconstruction. Compared to the existing methods based on mesh representations, these implicit function-based methods~\cite{karunratanakul2021skeleton, corona2022lisa} can produce fine-grained hand geometry in a resolution-independent manner.


\noindent \textbf{Interacting Hand-Object Reconstruction.}
Similar to single-hand reconstruction, most of the existing methods~\cite{hasson2019learning, hasson2020leveraging, baek2020weakly, doosti2020hope, rhoi2020} on hand-object reconstruction adopt MANO topology-based mesh representations to model a hand shape. Few recent works~\cite{karunratanakul2020grasping, karunratanakul2021skeleton} consider neural implicit representations to model hand-objects, but they often constrain their interaction based on contacts -- which does not \emph{necessarily} occur in hand-to-hand interactions~\cite{zhang2021interacting}. In addition, these methods mainly consider a rigid object in the interaction, which makes them ineffective to be directly applied for articulated two-hand reconstruction.

\noindent \textbf{Interacting Two-Hand Reconstruction.}
Compared to single-hand or hand-object reconstruction, two-hand reconstruction is more challenging, as more complex occlusions and deformations occur from the interaction between two articulated hands. For modeling two interacting hands, there have been methods recently proposed for two-hand \emph{pose} reconstruction from RGB images~\cite{moon2020interhand2, kim2021end, fan2021learning, li2022interacting, zhang2021interacting}, depth images~\cite{mueller2019real, taylor2017articulated}, or RGB-D video sequences~\cite{kyriazis2014scalable, oikonomidis2012tracking}. Yet, there are only a few methods that can directly reconstruct the \emph{dense surface} of closely interacting two-hands~\cite{li2022interacting, zhang2021interacting, rong2021monocular, mueller2019real}, which is more challenging and important in reasoning hand-to-hand interactions. Mueller \emph{et al.}~\cite{mueller2019real} propose an energy minimization framework to fit MANO~\cite{romero2017embodied} parameters to the input depth image of two interacting hands. Two-Hand-Shape-Pose~\cite{zhang2021interacting} introduce pose-aware attention and context-aware cascaded refinement modules to directly regress two-hand MANO parameters from an RGB image. IntagHand~\cite{li2022interacting} propose an attention-based graph convolutional network (GCN) for two-hand vertex regression from an RGB image.
These recent deep learning-based frameworks~\cite{zhang2021interacting, li2022interacting, rong2021monocular, wang2020rgb2hands} have shown that (1) an attention mechanism to model non-local interactions and (2) context-aware shape refinement steps are effective in two-hand reconstruction, which have inspired the design of our two-hand occupancy function. However, compared to these existing methods, our occupancy-based method can learn resolution-independent hand surface with better image-shape alignment, as our output space (i.e., occupancy field) itself is more directly aligned with the input image space.

\begin{figure*}[!t]
\begin{center}
\includegraphics[width=0.98\textwidth, height=0.48\textwidth]{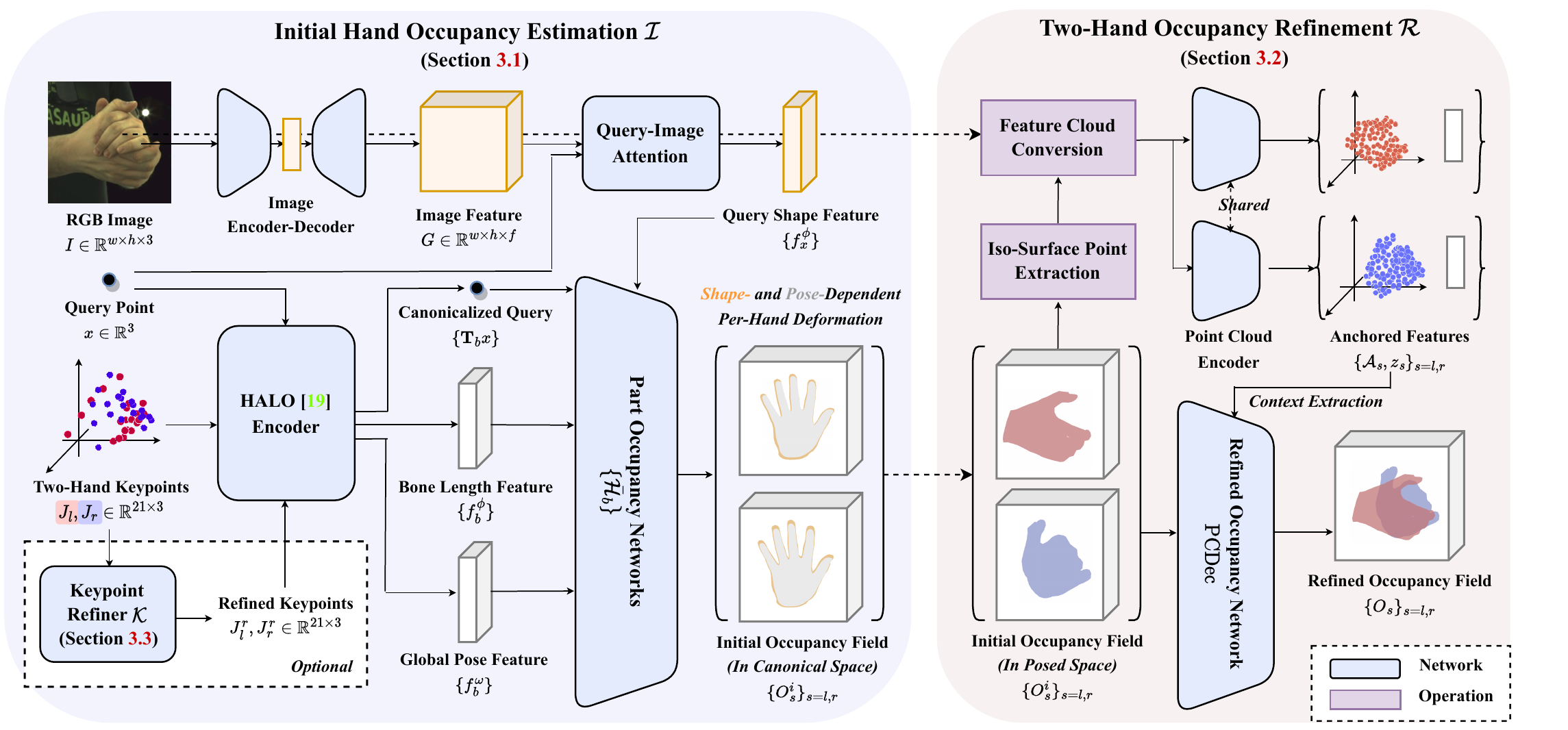}
\end{center}
\vspace{-\baselineskip}
\caption{\textbf{Architecture overview.} Given an RGB image and coarse 3D two-hand keypoints, our method estimates two-hand occupancy volumes via (1) initial hand occupancy estimation and (2) two-hand occupancy refinement. In a nutshell, the initial hand occupancy network uses query-image attention and HALO~\cite{karunratanakul2021skeleton} encoder to learn per-hand occupancy volumes in the canonical spaces. Then, the two-hand occupancy refinement network encodes the initial per-hand occupancy volumes into anchored features, which are then used to provide two-hand context information for refined occupancy estimation in the original posed space.}
\label{fig:architecture_overview}
\vspace{-\baselineskip}
\end{figure*}


\noindent \textbf{Neural Articulated Implicit Representation.}
Many existing methods to model articulated objects (e.g., human bodies, hands) use neural \emph{articulated} implicit representations~\cite{deng2020nasa, karunratanakul2021skeleton, corona2022lisa, noguchi2021neural, saito2021scanimate, mihajlovic2022coap}, which directly condition an implicit geometry on the object articulation.
While many methods~\cite{deng2020nasa, noguchi2021neural, saito2021scanimate, mihajlovic2022coap} mainly demonstrate their results on human body modeling, there are only a few studies that explore the effectiveness of implicit representations for hands. LISA~\cite{corona2022lisa} proposes an articulated VolSDF~\cite{yariv2021volume} to model shape and appearance of single-hands, but it assumes that various input domains (i.e., multi-view RGB image sequences, 3D bone transformations and foreground masks) are available and requires model optimization for inference. HALO~\cite{karunratanakul2021skeleton} proposes a single-hand articulated occupancy function driven by 3D keypoints via incorporating a novel canonicalization layer -- eliminating the need for the ground truth 3D bone transformations used in most of the articulated implicit functions~\cite{deng2020nasa, corona2022lisa, noguchi2021neural, saito2021scanimate, mihajlovic2022coap}. However, HALO models pose-independent shape variations only with hand bone lengths, which may be a strong assumption. We take inspiration from HALO to condition our two-hand occupancy on the pose represented as sparse 3D keypoints, but we further improve HALO by conditioning shape variations using an RGB image, which is a more direct form of shape observation than bone lengths.

\section{Im2Hands: Implicit Two-Hand Function}
\label{sec:method}


Im2Hands is a neural occupancy representation of two interacting hands. Our two-hand occupancy network can be formally defined as:

\vspace{-0.1\baselineskip}
\begin{equation}
\mathcal{O} (x\, |\, \alpha,\, \beta) \rightarrow [o_l,\, o_r],
\end{equation}
\vspace{-0.4\baselineskip}

\noindent where $\mathcal{O}$ is a neural network with the learned weights. $\mathcal{O}$ maps an input 3D query point $x \in \mathbb{R}^3$ to occupancy probabilities for each side of hand $o_l,\, o_r \in [0, 1]$ conditioned on a shape observation $\alpha$ and a pose observation $\beta$, which are represented as an RGB image $I \in \mathbb{R}^{w \times h \times 3}$ and sparse 3D two-hand keypoints $J = \left[ {J_l \atop J_r} \right]
 \in \mathbb{R}^{42 \times 3}$, respectively. 

One straightforward approach to design a neural implicit function for two hands would be directly applying the existing articulated occupancy function for \emph{single hand} (i.e., HALO~\cite{karunratanakul2021skeleton}) to both hands separately. This existing method can robustly model pose-dependent deformations via learning hand shapes in the canonical space, but it does not effectively capture other shape-dependent deformations (e.g.,  identity-dependent or soft tissue deformations). We thus design our initial hand occupancy estimation network (Section~\ref{subsec:initial_hand_occupancy_estimation}) that combines (1) HALO to robustly model \emph{pose-dependent deformations} by learning in the hand canonical space and (2) our novel query-image attention module to capture \emph{shape-dependent deformations} observed from an RGB image. However, it is difficult to handle two-hand interaction contexts while learning in the hand canonical spaces. Thus, we additionally propose a two-hand occupancy refinement network (Section~\ref{subsec:two-hand_occupancy_refinement}) to perform context-aware shape refinement of interacting two hands in the original posed space. Furthermore, we consider two-hand reconstruction from single images, where the input two-hand keypoints can be obtained by applying an off-the-shelf two-hand pose estimation method (e.g.,\cite{fan2021learning, moon2020interhand2, kim2021end, li2022interacting, zhang2021interacting}). To this end, we additionally introduce an optional input keypoint refinement module (Section~\ref{subsec:input_keypoint_refinement}) to alleviate input keypoint errors to enable more robust shape estimation.

In what follows, we explain each of the proposed components of Im2Hands in more detail.

\subsection{Initial Hand Occupancy Estimation}
\label{subsec:initial_hand_occupancy_estimation}

Given a 3D query point $x$, our initial hand occupancy network $\mathcal{I}$ predicts initial occupancy probabilities for each hand $o^{i}_l,\, o^{i}_r \in [0,\, 1]$ conditioned on an RGB image $I$ and two-hand keypoints $J$. Our network design is partly based on HALO~\cite{karunratanakul2021skeleton}, which models an implicit \emph{single-hand} conditioned on 3D keypoints. 

\vspace{0.5\baselineskip}
\noindent \textbf{Background: HALO~\cite{karunratanakul2021skeleton}.} HALO aims to learn an occupancy field of a single hand driven by 3D keypoints. To learn hand shapes in the canonical space, HALO proposes a novel algorithm to transform the input 3D keypoints to canonicalization matrices $\{\mathbf{T}_b\}_{\mathit{b=1}}^{B}$, where $\mathbf{T}_b$ is a transformation matrix
to the canonical pose for hand bone $b$, and $B$ is the number of hand bones. Using the obtained canonicalization matrices, hand occupancy at query $x$ is modeled as:

\vspace{-0.5\baselineskip}
\begin{equation}
\mathcal{H}(x\,|\, J) = \max\limits_{b = 1,\, ...,\, B}\{\bar{\mathcal{H}}_b(\mathbf{T}_bx,\, f_b^\phi, f_b^\omega)\},
\label{eq:halo}
\end{equation}

\noindent where $\bar{\mathcal{H}}_b$ is an MLP-based part occupancy network to learn the shape of hand bone $b$. Each part occupancy network takes a canonicalized query point $\mathbf{T}_{b}x$ along with a bone length shape feature $f_b^\phi$ and a pose feature $f_b^\omega$ for bone $b$. $f_b^\phi$ is extracted by an MLP encoder that takes a bone length vector $l \in \mathbb{R}^B$ computed from the input 3D keypoints. $f_b^\omega$ is obtained by another MLP encoder that takes global pose matrices $\{\mathbf{T}_{b}\}_{b=1}^B$ and a root translation vector $t \in \mathbb{R}^3$ as inputs. For more details about HALO, we kindly refer the reader to \cite{karunratanakul2021skeleton}.

\vspace{0.5\baselineskip}
\noindent \textbf{Our Initial Hand Occupancy Network.} While HALO is effective in modeling pose-dependent deformations via query point canonicalization (i.e. $\mathbf{T}_bx$) and pose feature extraction (i.e., $f^\omega_b$), its bone length shape feature (i.e., $f^\phi_b$) cannot model shape-dependent deformations that cannot be described by bone lengths. Modeling such shape variations is especially important in interacting two-hand reconstruction, as soft tissue deformations commonly occur due to inter-hand contacts. Thus, we introduce an additional shape feature conditioned on an RGB image~$I$, which provides a minimal observation for such shape-dependent deformations. We propose a per-query shape feature $f_x^\phi$ conditioned on the image $I$ as follows:



\begin{equation}
f_x^\phi = \mathrm{MSA}([\mathrm{PosEnc}(x),\, \mathrm{ImgEnc}(I)]),
\label{eq:query-image-att}
\vspace{0.3\baselineskip}
\end{equation}

\noindent where $\mathrm{MSA}$ denotes a multi-headed self-attention module that takes a query feature $\mathrm{PosEnc}(x)$ and patch-wise image features $\mathrm{ImgEnc}(I)$. To be more specific, $\mathrm{PosEnc}$ is an MLP that performs positional embedding to map the input query coordinate $x$ to a query feature vector $f_x \in \mathbb{R}^{d}$. $\mathrm{ImgEnc}$ is an image encoder of Vision Transformer~\cite{dosovitskiy2020image}, which maps the input image $I$ into patch-wise image features $f_I \in \mathbb{R}^{p \times d}$, where $p$ is the number of patches. Through this query-image attention, we can obtain a per-query shape feature $f_x^\phi$ contributed by the features of more \emph{relevant} image patch regions to the input query $x$. In Section~\ref{sec:experiments}, we also show that using this query-image attention yields better results than directly using a local image feature located at the projected query position (e.g., PIFu~\cite{saito2019pifu}).

Finally, our initial per-hand occupancy is modeled by feeding our per-query shape feature $f_x^\phi$ along with the other inputs to the MLP-based part occupancy network $\bar{\mathcal{H}}_b$ in Equation~\ref{eq:halo}:

\vspace{-0.6\baselineskip}
\begin{equation}
\mathcal{I}(x\,|\, I,\, J) = \max\limits_{b = 1,\, ...,\, B}\{\bar{\mathcal{H}}_b(\mathbf{T}_bx,\, f_b^\phi, f_x^\phi, f_b^\omega)\}.
\end{equation}

\noindent Note that our occupancy is estimated for each hand separately in this stage. We use the shared network for both hands, but we distinguish each hand by feeding coarse 3D keypoints of each hand $J_{l}, J_{r} \in \mathbb{R}^{21 \times 3}$ separately as inputs to our network. As these keypoints are used to compute the canonicalization matrices and the pose feature, we can robustly learn our initial occupancy in the canonical space defined per-hand. 



\subsection{Two-Hand Occupancy Refinement}
\label{subsec:two-hand_occupancy_refinement}

The previous step can robustly estimate per-hand occupancy volumes, but it does not effectively model the coherency between two hands due to the adaptation of the hand canonical spaces. As recent mesh-based two-hand reconstruction methods~\cite{zhang2021interacting, li2022interacting} have shown that (1) two hand shapes are correlated with each other and thus (2) context-aware two-hand shape refinement is effective, we additionally propose a two-hand occupancy refinement network $\mathcal{R}$ that refines the initial two-hand shape \emph{in the original posed space}. It estimates the refined two-hand occupancies $o_l,\, o_r \in [0, 1]$ at the query point $x$ conditioned on (1) the initial two-hand occupancy probabilities at $x$ estimated by $\mathcal{I}$ and (2) the input image $I$.

To condition our refined occupancy estimation on the interaction context, it is necessary to encode the initial geometry of two hands. To effectively learn features from the current two hand shapes, we first represent them as point clouds $\mathcal{P}_{l}, \mathcal{P}_{r} \in \mathbb{R}^{n \times 3}$, which are sets of iso-surface points of each side of hand geometry. These points can be easily obtained by collecting 512 farthest points among the query coordinates evaluated to be on surface by our initial hand occupancy network: $\{ x\, |\, 0.5 - \epsilon \leq \mathcal{I}(x\, |\, I, J) \leq 0.5 + \epsilon\}$. We then encode each hand point cloud into a global latent vector and local latent vectors anchored in 3D space:

\begin{equation}
\{z_{s}, \mathcal{A}_{s} = \mathrm{PCEnc}(\mathcal{P}_{s}, I)\}_{s = l, r},
\label{eq:pcd_encoding}
\end{equation}

\noindent where $z_{s} \in \mathbb{R}^{z}$ is a global latent vector and $\mathcal{A}_{s} \in \mathbb{R}^{m\times(3+a)}$ is a set of $m$ number of anchor points (i.e., a subset of the input point cloud selected by farthest point sampling) with $a$-dimensional per-point features. In our point cloud encoding module $\mathrm{PCEnc}$, we first preprocess the input point cloud $\mathcal{P}_{s}$ into a feature cloud $\mathcal{F}_{s} \in \mathbb{R}^{n \times (3 + f)}$ to incorporate texture information observed from the input image $I$. To this end, we apply a simple encoder-decoder CNN to extract an image feature map $G \in \mathbb{R}^{w \times h \times f}$ and concatenate each point $p$ in $\mathcal{P}_s$ to an image feature vector located at the projected position of $p$ in $G$. We then feed our feature cloud to the encoder of AIR-Net~\cite{giebenhain2021air} that extracts global and locally anchored point cloud features using Point Transformer~\cite{zhao2021point}.

Using the obtained point cloud features, we extract a latent code that encodes interacting two-hand shapes and image context as follows:

\begin{equation}
z_{c} = \mathrm{ContextEnc}(z_{l}, z_{r}, z_{I}).
\label{eq:context_encoding}
\end{equation}

\noindent $\mathrm{ContextEnc}$ is an MLP that extracts the context feature, and $z_{I}$ is the global bottleneck feature from the image encoder-decoder previously used to extract $G$. Finally, we estimate our refined occupancy at the query point $x$ as follows:

\vspace{-0.5\baselineskip}
\begin{equation}
\{o_{s} = \mathrm{PCDec}(x, o^i_s, \mathcal{A}_{s},\, z_{c}) \}_{s = l, r},
\label{eq:pcd_decoding}
\end{equation}

\noindent where $\mathrm{PCDec}$ is a point cloud decoder, for which we adopt a similar architecture to the decoder of AIR-Net~\cite{giebenhain2021air}. The original AIR-Net decoder predicts an occupancy probability given a single point cloud via vector cross attention between the query point $x$ and local anchor features $\mathcal{A}_s$ and a global feature $z_s$ (please refer to \cite{giebenhain2021air} for more details). Our decoder instead (1) uses a global latent vector $z_c$ that encodes the context between the input image and the initial two hand shapes predicted by $\mathcal{I}$ and (2) conditions the occupancy estimation on our initial occupancy probability by concatenating $o^i_s$ at $x$ to the input query coordinate. In summary, our refined occupancy estimation is effectively conditioned both on (1) local latent descriptors $\mathcal{A}_{s}$ that encode information about the initial hand geometry, (2) a global latent descriptor $z_c$ that encodes the global context of two-hand interaction and the input image, and (3) our initial occupancy estimation $o^i_s$. 

\subsection{Input Keypoint Refinement}
\label{subsec:input_keypoint_refinement}

In this section, we further consider image-based two-hand reconstruction using Im2Hands, where no ground truth hand keypoints are available. To enable robust shape reconstruction from keypoints \emph{predicted} from an off-the-shelf image-based two-hand keypoint estimator (e.g., \cite{moon2020interhand2, kim2021end, fan2021learning, li2022interacting, zhang2021interacting}), we introduce an optional keypoint refinement module $\mathcal{K}$ that can alleviate noise in the input two-hand keypoints. Our keypoint refinement module is formulized as: $\mathcal{K}(J, I) \rightarrow J^r$, where $J, J^r \in \mathbb{R}^{42 \times 3}$ are initial and refined 3D two-hand keypoints, respectively. To be more specific, $\mathcal{K}$ is designed as:  

\vspace{-0.8\baselineskip}
\begin{equation}
\mathcal{K}(J, I) = \mathrm{MSA}([\mathrm{GCN}(\mathrm{KptEnc}(J)),\, \mathrm{ImgEnc}(I)]).
\end{equation}
\vspace{-0.8\baselineskip}

\noindent In the above equation, we first extract input keypoint features by $\mathrm{KptEnc}$ that encodes the index and coordinate of each keypoint using a shared MLP. We then build a two-hand skeleton graph with initial node features set as features extracted from the previous $\mathrm{KptEnc}$ function. Next, we feed this initial two-hand skeleton graph to a $\mathrm{GCN}$ to extract keypoint features that further encode information of the initial two-hand structure. Finally, we apply multi-headed self-attention ($\mathrm{MSA}$) between the keypoint features and image features -- similarly to query-image attention in Equation~\ref{eq:query-image-att} -- to regress the refined two-hand keypoint positions. 
In Section~\ref{sec:experiments}, we experimentally demonstrate that $\mathcal{K}$ significantly helps improving the quality of two-hand reconstruction from single images. 


\subsection{Loss Functions}
\label{subsec:loss_functions}

We now explain loss functions to train our two-hand occupancy network. 

\noindent \textbf{Initial Hand Occupancy Network ($\mathcal{I}$).} $\mathcal{I}$ is trained by MSE loss that measures the deviation between the ground truth and the predicted occupancy probabilities. For training query point generation, we combine (1) points uniformly sampled in the hand bounding box and (2) points sampled on the hand surface added with Gaussian noise following~\cite{karunratanakul2021skeleton, saito2019pifu}.

\noindent \textbf{Two-Hand Occupancy Refinement Network ($\mathcal{R}$).} Similar to $\mathcal{I}$, $\mathcal{R}$ is trained by MSE loss using the ground truth occupancy supervision. It is additionally trained with penetration loss, which penalizes the refined two-hand occupancy values that are estimated to be occupied in both hands at the same query position. Given a set of training samples $\mathcal{X}$, a set of sampled training query points $\mathcal{P}$, and $\mathcal{R}_l$ and $\mathcal{R}_r$ that output the refined occupancy probabilities for left and right hands, our penetration loss is defined as:

\vspace{-0.6\baselineskip}
\begin{equation}
\mathcal{L}_{pen} = \frac{1}{|\mathcal{X}|}\sum_{(I, J) \in \mathcal{X}}\sum_{x \in \mathcal{P}}\mathcal{R}_{l}(x | I,\, J) \cdot \mathcal{R}_{r}(x | I,\, J),
\end{equation}

\noindent where $\mathcal{R}_l(\cdot) > 0.5$ and $\mathcal{R}_r(\cdot) > 0.5$. We incorporate this loss function to avoid inter-penetration between two hands.

\noindent \textbf{Input Keypoint Refinement Network ($\mathcal{K}$).} $\mathcal{K}$ is trained by MSE loss computed using the ground truth and the predicted two-hand keypoints. Due to the space limit, please refer to our supplementary section for training and network architecture details.

\begin{figure*}[!t]
\begin{center}
\includegraphics[width=\textwidth]{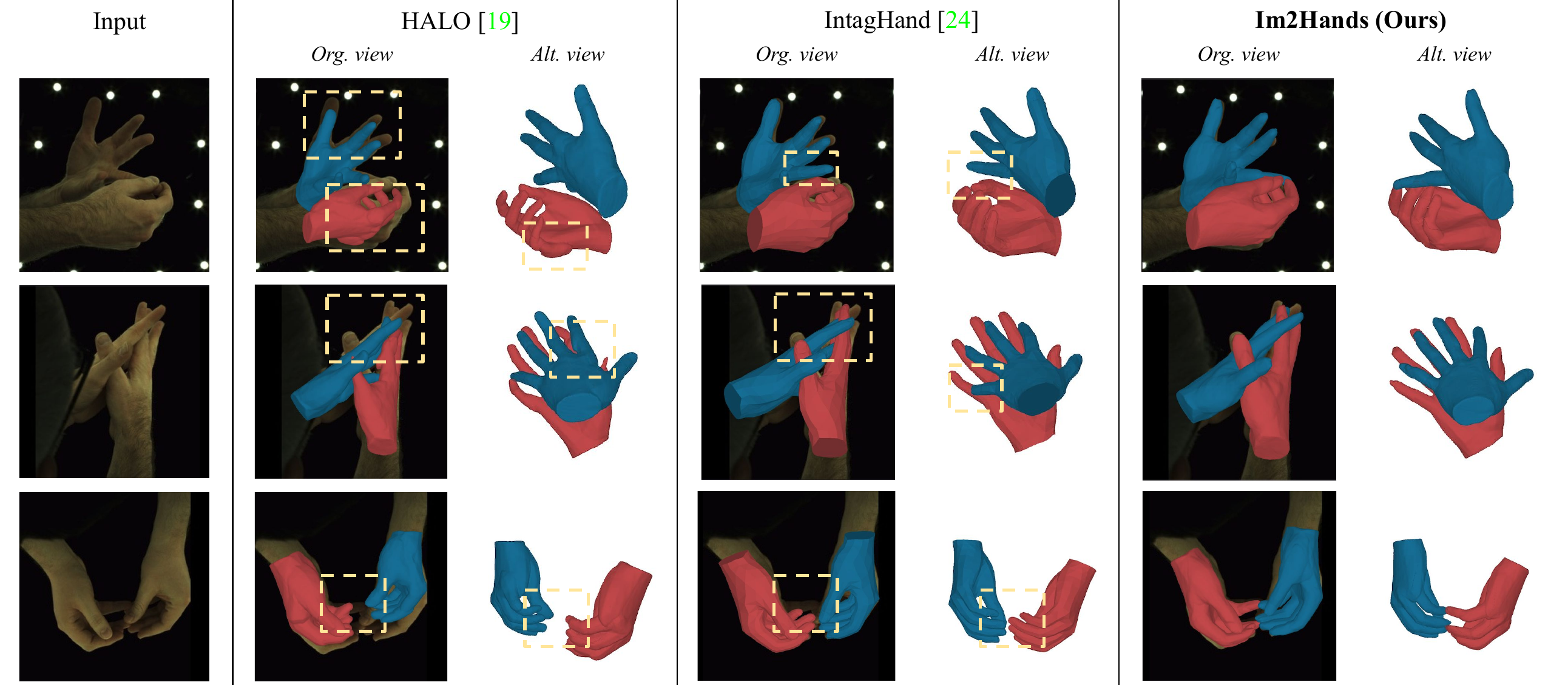} 
\caption{\textbf{Qualitative comparison of \emph{single-image} interacting two-hand reconstruction on InterHand2.6M~\cite{moon2020interhand2}.} We compare our results to HALO~\cite{karunratanakul2021skeleton} and IntagHand~\cite{li2022interacting}, where our method produces two-hand shapes with better hand-to-image and hand-to-hand alignment. Ours also properly captures finger
contacts (\emph{rows 1 and 3}) and avoids shape penetration (\emph{row 2}). For our method and HALO, we use the hand keypoints predicted by DIGIT~\cite{fan2021learning} to condition the articulated occupancy fields. Please see the supplementary section for more qualitative examples.} 
\label{fig:main_qualitative_results}
\vspace{-1.4\baselineskip}
\end{center}
\end{figure*}

\section{Experiments}
\label{sec:experiments}


\subsection{Experimental Setups}
\label{subsec:experimental_setting}

\subsubsection{Datasets, Metrics, and Keypoint Estimatiors}

\noindent \textbf{Datasets.}
We mainly use InterHand2.6M~\cite{moon2020interhand2} dataset -- the only interacting two-hand dataset with dense shape annotations -- for both quantitative and qualitative evaluation. To maintain consistency with the previous work~\cite{li2022interacting}, we only use interacting hand (IH) samples annotated as \emph{valid} hand type. The resulting dataset contains 366K training samples, 110K validation samples, and 261K test samples. For qualitative evaluation, we additionally demonstrate our results on RGB2Hands~\cite{wang2020rgb2hands} and EgoHands~\cite{bambach2015lending} datasets, which contain RGB videos of interacting two hands without shape annotations.

\noindent \textbf{Evaluation Metrics.}
For evaluating the quality of reconstructed two-hand shapes, we compute the mean Intersection over Union (IoU) and Chamfer L1-Distance (CD) between the predicted and the ground truth two-hand shapes. We also evaluate the accuracy of 3D hand keypoints after the proposed keypoint refinement using Mean Per Joint Position Error (MPJPE).

\noindent \textbf{Two-Hand Keypoint Estimation Methods.} To evaluate Im2Hands on \emph{single-image} two-hand reconstruction, we leverage an off-the-shelf image-based two-hand keypoint estimation method. We consider DIGIT~\cite{fan2021learning} and IntagHand~\cite{li2022interacting}, whose official implementation is publicly available. We would like to note that IntagHand is also proposed for two-hand \emph{shape} estimation, thus we additionally consider it as a baseline for two-hand shape reconstruction. However, when using IntagHand for a \emph{keypoint} estimation method in our image-based reconstruction experiments, we completely discard the predicted shape information and only use the estimated keypoints -- considering it as a pure two-hand keypoint estimation method. We emphasize that our two-hand occupancy function works agnostic to the architecture of the previous two-hand keypoint pre-processor and can be used in a plug-and-play manner.

\subsubsection{Baselines}
\label{subsubsection:baselines}
As Im2Hands is the first neural implicit function for two interacting hands, we compare our method to (1) the existing \emph{single-hand} reconstruction method using \emph{implicit} representation~\cite{karunratanakul2021skeleton} and (2) \emph{two-hand} reconstruction methods using \emph{mesh} representation~\cite{li2022interacting, zhang2021interacting}.

\vspace{0.5\baselineskip}
\noindent \textbf{Implicit Single-Hand Reconstruction Methods.}
We consider HALO~\cite{karunratanakul2021skeleton}, which is a neural implicit single-hand representation driven by 3D keypoints. As HALO models a hand shape only using 3D keypoints, we modify HALO to additionally leverage an input RGB image by feeding pixel-aligned features~\cite{saito2019pifu} together with the other inputs to the part occupancy functions -- to make more fair comparison to Im2Hands by using the same input data domains (see Section~\ref{subsec:representation_power}). Although LISA~\cite{corona2022lisa} is another implicit single-hand representation which takes both RGB image and 3D keypoints, we do not make direct comparison to LISA, as it requires more input data domains (e.g., foreground masks, multi-view RGB images) than Im2Hands.

\vspace{0.5\baselineskip}
\noindent \textbf{Mesh-Based Two-Hand Reconstruction Methods.}
We compare our method to IntagHand~\cite{li2022interacting} and Two-Hand-Shape-Pose~\cite{zhang2021interacting}, which are the state-of-the-art methods on interacting two-hand shape reconstruction. These methods reconstruct a fixed-topology mesh from an RGB image by using dense vertex correspondence~\cite{li2022interacting} or statistical model parameter annotations~\cite{zhang2021interacting} for training. Since they are designed for \emph{single-image} two-hand reconstruction and do not take 3D hand keypoints as input, we also evaluate our model on image-based reconstruction by using 3D hand keypoints predicted from an image (see Section~\ref{subsec:single_image_reconstruction}).


\subsection{Reconstruction From Images and Keypoints}
\label{subsec:representation_power}

In this section, we examine the representation power of Im2Hands given an input RGB image paired with the ground truth 3D hand keypoints (e.g., obtained from a sensor). In Table~\ref{table:representation_power}, we quantitatively compare our reconstruction results to the baseline methods. In the fifth row, note that we also compare our method to a re-implimented version of HALO~\cite{karunratanakul2021skeleton} that takes the same input data domains as ours (please refer to Section~\ref{subsubsection:baselines}). In the table, Im2Hands outperforms all the other methods by effectively leveraging an RGB image and hand keypoints to model shape- and pose-dependent deformations, respectively.


\begin{table}[!h]
\centering
{ \small
\setlength{\tabcolsep}{0.2em}
\renewcommand{\arraystretch}{1.0}
\caption{\textbf{Comparison of interacting two-hand reconstruction on InterHand2.6M~\cite{moon2020interhand2}.} For input domains of each method, $\mathbf{I}$, $\mathbf{J}$, and $\mathbf{L}$ denote RGB image, 3D hand keypoints, and hand bone lengths, respectively.}
\label{table:representation_power}

\begin{tabularx}{\columnwidth}{>{\centering}m{3.6cm}|Y|Y|Y}
\toprule
Method & Inputs & IoU \footnotesize{(\%)} $\uparrow$ & CD \footnotesize{(mm)} $\downarrow$ \\
\midrule
Two-Hand-Shape-Pose~\cite{zhang2021interacting} & $\mathbf{I}$, $\mathbf{L}$ & 54.8 & 5.51 \\
IntagHand~\cite{li2022interacting} & $\mathbf{I}$, $\mathbf{L}$ & 67.0 & 3.88 \\
\midrule
HALO~\cite{karunratanakul2021skeleton} & $\mathbf{J}$ & 74.7 & 2.62 \\ 
\midrule
HALO (modified)~\cite{karunratanakul2021skeleton} & $\mathbf{I}$, $\mathbf{J}$ & 75.8 & 2.51 \\
\midrule
\textbf{Im2Hands (Ours)} & $\mathbf{I}$, $\mathbf{J}$ & \textbf{77.8} & \textbf{2.30} \\
\bottomrule
\end{tabularx}
}
\vspace{-0.5\baselineskip}
\end{table}

\subsection{Reconstruction From Single Images}
\label{subsec:single_image_reconstruction}

In this section, we evaluate Im2Hands on single-image two-hand reconstruction. 
To this end, we estimate two-hand shapes using 3D hand keypoints predicted from an input image via off-the-shelf two-hand keypoint estimation methods~\cite{li2022interacting, fan2021learning} and the proposed keypoint refinement module ($\mathcal{K}$). Note that the goal of this experiment is to evaluate each method in a setting where \emph{no ground truth 3D hand keypoints} are available. Thus, we disabled the rescaling of the reconstructed hand joints and shapes performed by some of the existing methods~\cite{li2022interacting, zhang2021interacting} using the scale ratio calculated from a subset of the \emph{ground truth} 3D keypoints at the test time. For those methods~\cite{li2022interacting, zhang2021interacting} that require such rescaling, we use the mean scale of the hands in the training set of InterHand2.6M~\cite{moon2020interhand2} to perform fair comparison.

In Table~\ref{table:joint_results}, we first evaluate the effectiveness of our keypoint refinement module on noisy two-hand keypoints predicted by DIGIT \cite{fan2021learning} and IntagHand \cite{li2022interacting}. Our method is successful in alleviating input keypoint errors. Especially, it is shown to be effective when the degree of the input keypoint error is high.

\begin{table}[!h]
\centering
{ \small
\setlength{\tabcolsep}{0.2em}
\renewcommand{\arraystretch}{1.0}
\caption{\textbf{Effectiveness of our keypoint refinement module ($\mathcal{K}$) on InterHand2.6M~\cite{moon2020interhand2}.} We use two-hand keypoints predicted by DIGIT~\cite{fan2021learning} and IntagHand~\cite{li2022interacting} as inputs to our method.}
\label{table:joint_results}

\begin{tabularx}{\columnwidth}{>{\centering}m{4.5cm}|Y}
\toprule
Method & MPJPE \footnotesize{(mm)} $\downarrow$\\
\midrule
DIGIT~\cite{fan2021learning} & 16.75 \\
\textbf{DIGIT~\cite{fan2021learning} + $\mathcal{K}$ (Ours)} & \textbf{10.70} \\
\midrule
IntagHand~\cite{li2022interacting} & 10.13 \\ 
\textbf{IntagHand~\cite{li2022interacting} + $\mathcal{K}$ (Ours)} & \textbf{9.68} \\ 
\bottomrule
\end{tabularx}
}
\vspace{-0.5\baselineskip}
\end{table}

We now report our two-hand shape reconstruction results from single images. In Table~\ref{table:image_results}, our method achieves the state-of-the-art results in image-based interacting two-hand reconstruction on InterHand2.6M~\cite{moon2020interhand2}. We would like to re-emphasize that the baselines originally designed for image-based two-hand reconstruction~\cite{li2022interacting, zhang2021interacting} (1) can produce only fixed-resolution meshes and (2) require dense vertex correspondences or statistical model parameter annotations for training, while our method outperforms them without such constraints. Also, note that most of the existing articulated implicit functions (e.g., \cite{karunratanakul2021skeleton, deng2020nasa, corona2022lisa}) are typically evaluated using \emph{noiseless} keypoints or skeletons. In contrast, our method is experimentally demonstrated to achieve competitive results in a setting where \emph{no ground truth keypoints} are available -- by leveraging the proposed
keypoint and two-hand shape refinement modules.

\begin{table}[!h]
\centering
{ \small
\setlength{\tabcolsep}{0.2em}
\renewcommand{\arraystretch}{1.0}
\caption{\textbf{Comparison of \emph{single-image} interacting two-hand reconstruction on InterHand2.6M~\cite{moon2020interhand2}.} We use two-hand keypoints predicted by DIGIT~\cite{fan2021learning} and IntagHand~\cite{li2022interacting} as inputs to our method and HALO~\cite{karunratanakul2021skeleton}.}
\label{table:image_results}

\begin{tabularx}{\columnwidth}{>{\centering}m{5cm}|Y|Y}
\toprule
Method & IoU \footnotesize{(\%)} $\uparrow$ & CD \footnotesize{(mm)}$\downarrow$ \\
\midrule
Two-Hand-Shape-Pose~\cite{zhang2021interacting} & 48.4 & 6.09 \\
IntagHand~\cite{li2022interacting} & 59.0 & 4.69 \\
\midrule
DIGIT~\cite{fan2021learning} + HALO~\cite{karunratanakul2021skeleton} & 45.1 & 7.64 \\
IntagHand~\cite{li2022interacting} + HALO~\cite{karunratanakul2021skeleton} & 53.8 & 5.38 \\
\midrule
\textbf{DIGIT~\cite{fan2021learning} + Im2Hands (Ours)} & \textbf{59.4} & \textbf{4.75} \\ 
\textbf{IntagHand~\cite{li2022interacting}+ Im2Hands (Ours)} & \textbf{62.1} & \textbf{4.35} \\ 
\bottomrule
\end{tabularx}
}
\vspace{-0.5\baselineskip}
\end{table}

Our qualitative comparison of image-based two-hand reconstruction is also shown in Figure~\ref{fig:main_qualitative_results}. Note that HALO~\cite{karunratanakul2021skeleton} produces non-smooth shapes due to the lack of shape or keypoint refinement steps given the noisy keypoint observation. IntagHand~\cite{li2022interacting} does not properly model finger contacts (\emph{rows 1 and 3}) or produces shape penetration (\emph{row 2}). Our method generates more plausible two-hand shapes that also align well with the input image.

\subsection{Generalizability Test}
\label{subsec:generalizability_test}

\begin{figure}[!h]
\begin{center}
\includegraphics[width=0.47\textwidth]{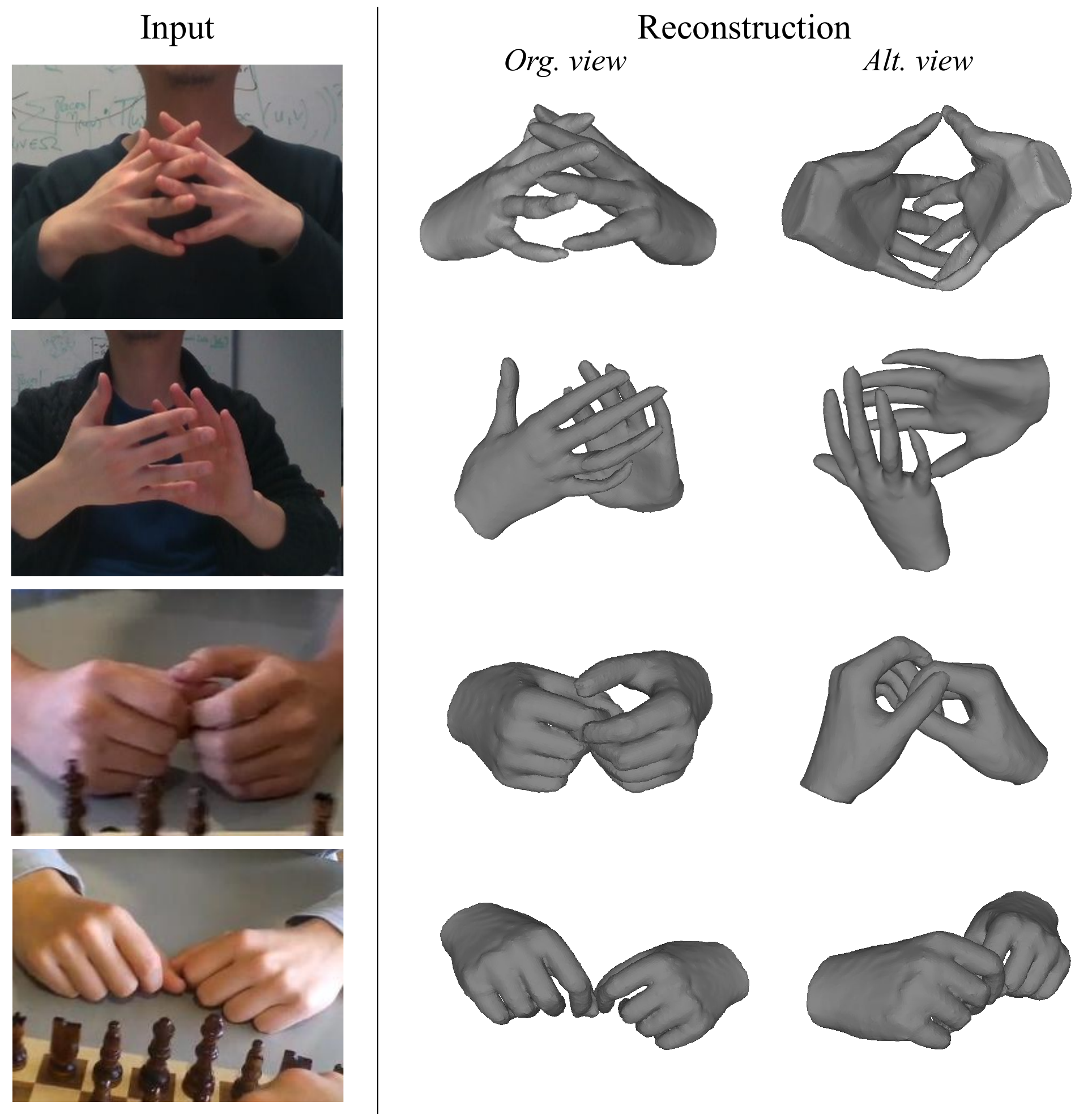} 
\caption{\textbf{Generalizability test on RGB2Hands~\cite{wang2020rgb2hands} and EgoHands~\cite{bambach2015lending}.} Top two rows show examples from RGB2Hands~\cite{wang2020rgb2hands} and bottom two rows show examples from EgoHands~\cite{bambach2015lending}.}
\label{fig:generalizability_test}
\vspace{-1.6\baselineskip}
\end{center}
\end{figure}

In Figure~\ref{fig:generalizability_test}, we additionally show our qualitative results on RGB2Hands~\cite{wang2020rgb2hands} and EgoHands~\cite{bambach2015lending} datasets. For two-hand keypoint estimation, we use DIGIT~\cite{fan2021learning} and IntagHand~\cite{li2022interacting} for RGB2Hands and EgoHands, respectively. Note that these reconstruction examples were directly generated by Im2Hands trained on InterHand2.6M~\cite{moon2020interhand2} dataset only, demonstrating the generalization ability of Im2Hands. For more detailed setups for this experiment, please refer to our supplementary section.

\subsection{Ablation Study}
\label{subsec:ablation_study}

In this section, we perform an ablation study to investigate the effectiveness of the major modules (i.e., $\mathcal{K}$, $\mathcal{I}$, and $\mathcal{R}$) of Im2Hands using 3D hand keypoints predicted by DIGIT~\cite{fan2021learning}. In Table~\ref{table:ablation_study_refinement}, our full model is the most effective compared to the other model variants. Please refer to the supplementary section for ablation study results with respect to more detailed model variations (e.g. models after removing \emph{each} of the proposed components inside $\mathcal{I}$ and $\mathcal{R}$).

\begin{table}[!h]
\centering
{ \small
\setlength{\tabcolsep}{0.2em}
\renewcommand{\arraystretch}{1.0}
\caption{\textbf{Ablation study using the \emph{predicted} 3D hand keypoints.} Experiments are performed on InterHand2.6M~\cite{moon2020interhand2} dataset using the keypoints predicted by DIGIT~\cite{fan2021learning}.}
\label{table:ablation_study_refinement}

\begin{tabularx}{\columnwidth}{>{\centering}m{4.8cm}|Y|Y}
\toprule
Method & IoU \footnotesize{(\%)} $\uparrow$ & CD \footnotesize{(mm)} $\downarrow$ \\
\midrule
$\mathcal{I}$ & 45.6 &  7.61 \\
$\mathcal{K}$ + $\mathcal{I}$ & 55.4 & 5.36\\ 
\midrule
\textbf{Im2Hands ($\mathcal{K}$ + $\mathcal{I}$ + $\mathcal{R}$)} & \textbf{59.4} & \textbf{4.75} \\ 
\bottomrule
\end{tabularx}
}
\vspace{-0.5\baselineskip}
\end{table}

\section{Conclusion and Future Work}

We present Im2Hands, the first neural implicit representation of two interacting hands. To effectively model the interaction context between two articulated hands, we propose two modules for initial occupancy estimation in the hand canonical space and context-aware occupancy refinement in the original posed space, respectively. Furthermore, we introduce an optional input keypoint refinement module to enable robust shape reconstruction from single images. As a result, we achieve two-hand shape reconstructions with better hand-to-image and hand-to-hand alignment compared to the baseline methods.

\vspace{0.5\baselineskip}
\noindent \textbf{Limitations and Future Work.}
 As Im2Hands models twohand articulated occupancy conditioned on the pose, the reconstruction accuracy depends on the quality of the input keypoints. We plan to further investigate the ways to effectively learn both pose and shape of two interacting hands in an end-to-end manner given a single image input.

\vspace{0.5\baselineskip}
\noindent \textbf{Acknowledgements.}
This work is in part sponsored by NST grant (CRC 21011, MSIT) and KOCCA grant (R2022020028, MCST). Minhyuk Sung acknowledges the support of the NRF grant (RS-2023-00209723) and IITP grant (2022-0-00594) funded by the Korean government (MSIT), and grants from Adobe, ETRI, KT, and Samsung Electronics.


{\small
\bibliographystyle{ieee_fullname}
\bibliography{egbib}
}

\newpage

  \makeatletter
  \newcommand{\manuallabel}[2]{\def\@currentlabel{#2}\label{#1}}
  \makeatother
  \manuallabel{subsec:generalizability_test}{4.4}
  \manuallabel{subsec:ablation_study}{4.5}
  \manuallabel{table:ablation_study_refinement}{4}
  \manuallabel{eq:query-image-att}{3}
  \manuallabel{eq:pcd_encoding}{5}
  \manuallabel{eq:context_encoding}{6}
  \manuallabel{eq:pcd_decoding}{7}
  \manuallabel{fig:supp_ablation}{S1}
  \manuallabel{fig:supp_comparison}{S2}
  \manuallabel{fig:supp_seq}{S3}
  \newcommand{\refpaper}[1]{in the paper}

\counterwithin{figure}{section}
\counterwithin{table}{section}

\renewcommand{\thesection}{S}
\renewcommand{\thetable}{S\arabic{table}}
\renewcommand{\thefigure}{S\arabic{figure}}

\vspace{-\baselineskip}
\clearpage
\section{Supplementary Material}

In this supplementary document, we first show more qualitative results of our method in Section~\ref{subsec:additional_qualitative_results}. We then show the results of our additional ablation study in Section~\ref{subsec:additional_ablation_study}. Finally, we report our implementation details and experimental details in Section~\ref{subsec:implementation_details} and \ref{subsec:detailed_experimental_setups}, respectively.

\subsection{Additional Qualitative Results}
\label{subsec:additional_qualitative_results}

\subsubsection{Video Results}
\vspace{-0.2\baselineskip}

We provide the video results (\url{https://youtu.be/3yNGSRz564A}) of our method on image-based two-hand reconstruction in comparison to HALO~\cite{karunratanakul2021skeleton} and IntagHand~\cite{li2022interacting}. This video contains the reconstruction results on InterHand2.6M~\cite{moon2020interhand2} test image sequences that are used in the main experiments in the paper. For our method and HALO, we use DIGIT~\cite{fan2021learning} to generate keypoint inputs from single images. Please note that our method and the baseline methods~\cite{li2022interacting, karunratanakul2021skeleton} are originally proposed for two-hand reconstruction from single images and/or keypoints, thus the shapes were reconstructed from \emph{each frame independently}. One important future research direction would be to extend our model to additionally utilize temporal information for tracking applications.

\vspace{-0.3\baselineskip}

\subsubsection{Ablation Study}
\vspace{-0.3\baselineskip}

In Figure~\ref{fig:supp_ablation}, we show the qualitative examples of our ablation study in Table~\ref{table:ablation_study_refinement} in the main paper. The shown examples are produced from single images, where we use the keypoints predicted by DIGIT~\cite{fan2021learning} as inputs. In the figure, $\mathcal{I}$ produces two-hand shapes that do not look plausible due to the input errors from the \emph{predicted} two-hand keypoints. $\mathcal{K + I}$ generates more plausible shapes through input keypoint refinement performed by $\mathcal{K}$, however, it still does not properly model hand-to-hand interactions (e.g., finger contacts). Our full model, $\mathcal{K + I + R}$, reconstructs the most accurate shapes with higher hand-to-image and hand-to-hand coherency.

\vspace{-0.3\baselineskip}

\subsubsection{Additional Qualitative Comparison}
\vspace{-0.3\baselineskip}

In Figure~\ref{fig:supp_comparison} (\emph{please see the next page}), we also show the additional examples of our qualitative comparison of interacting two-hand reconstruction on InterHand2.6M~\cite{moon2020interhand2}. Compared to HALO~\cite{karunratanakul2021skeleton} and IntagHand~\cite{li2022interacting}, Im2Hands can reconstruct interacting two-hand shapes with \textbf{a higher resolution, less penetrations, and better hand-to-image and hand-to-hand alignments}. The shown examples were produced from single image inputs to perform a fair comparison with IntagHand, where our method and HALO again leveraged DIGIT~\cite{fan2021learning} as a keypoint estimator.

\begin{figure}[!h]
\begin{center}
\includegraphics[width=0.5\textwidth]{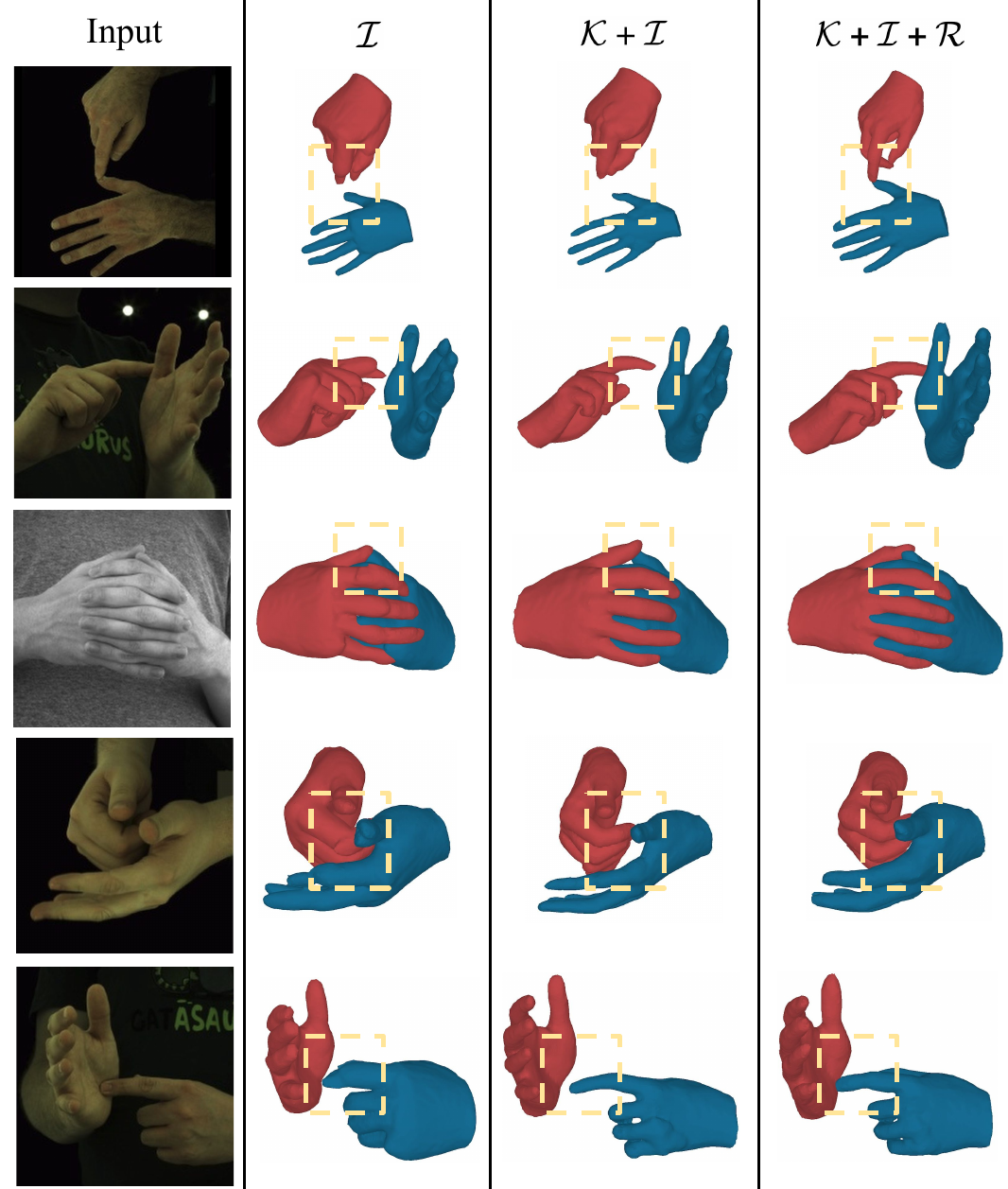}
\caption{\textbf{Qualitative examples of ablation study on InterHand2.6M~\cite{moon2020interhand2}.} $\mathcal{I}$, $\mathcal{R}$ and $\mathcal{K}$ denotes Initial Hand Occupancy Network, Two-Hand Occupancy Refinement Network, and Input Keypoint Refinement Network, respectively. }
\end{center}
\vspace{-\baselineskip}
\end{figure}

\begin{figure*}[!h]
\begin{center}
\includegraphics[width=\textwidth]{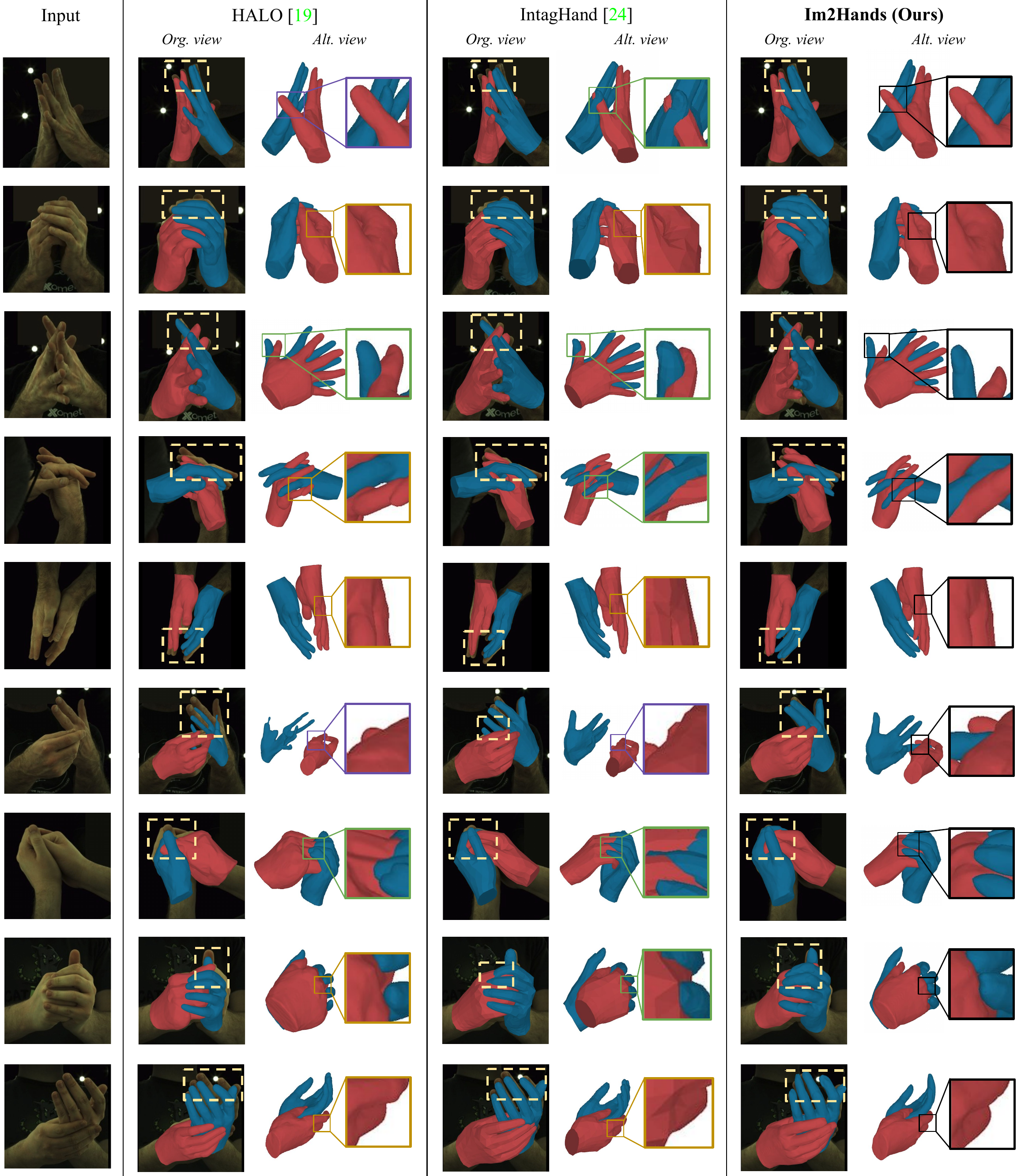}
\caption{\textbf{Additional qualitative examples of \emph{image-based} interacting two-hand reconstruction on InterHand2.6M~\cite{moon2020interhand2}.} We compare the results of our method with HALO~\cite{karunratanakul2021skeleton} and IntagHand~\cite{li2022interacting}. \textcolor[rgb]{0.0, 0.7, 0.0}{\textbf{Green boxes}} show penetrations, \textcolor[rgb]{0.8, 0.5, 0.2}{\textbf{brown boxes}} show non-smooth shapes, and \textcolor[rgb]{0.58, 0.44, 0.86}{\textbf{purple boxes}} show shapes with bad image alignment. Our method produces two-hand shapes with \textbf{better hand-to-image and hand-to-hand coherency, less penetrations, and a higher resolution}.}
\end{center}
\end{figure*}


\subsection{Additional Ablation Study}
\label{subsec:additional_ablation_study}

We now report the quantitative results of more detailed ablation study. In what follows, we first explain the notations for each of the evaluated variations of Im2Hands.


\begin{itemize}
    \item $\boldsymbol{\mathcal{I} - \mathrm{ImageCond}}$ denotes a variation where no image conditioning is used in $\mathcal{I}$, resulting in a model equivalent to HALO~\cite{karunratanakul2020grasping}.

    \item $\boldsymbol{\mathcal{I} - \mathrm{QueryImageAtt}}$ denotes a variation where no query-image attention is used in $\mathcal{I}$. Instead, pixel-aligned features (e.g., PIFu~\cite{saito2019pifu}) are used to condition our initial occupancy on an input image.

    \item $\boldsymbol{\mathcal{I} + \mathcal{R} - \mathrm{InitOccCond}}$ denotes a variation where the initial occupancy probability estimated by $\mathcal{I}$ is not used to condition our two-hand refined occupancy estimation in $\mathcal{R}$.

    \item $\boldsymbol{\mathcal{I} + \mathcal{R} - \mathrm{FeatureCloud}}$ denotes a variation where the feature cloud conversion is not performed in $\mathcal{R}$.

    \item $\boldsymbol{\mathcal{I} + \mathcal{R} - \mathrm{ContextLatent}}$ denotes a variation where the context latent extraction is not performed in $\mathcal{R}$. Instead, global latent vector of each hand point cloud is used in the refined occupancy estimation.

\end{itemize}

\noindent \textbf{Detailed Ablation Study With Ground Truth Keypoints.}
In Table~\ref{table:detailed_ablation_study}, our quantitative results across the variations of Im2Hands are shown. Note that the \emph{ground truth} keypoint inputs are used in these experiments. Our results demonstrate that each of the proposed model components contributes to more accurate two-hand shape estimation, and thus the proposed full model achieves the best performance.

\begin{table}[!h]
\centering
{ \small
\setlength{\tabcolsep}{0.2em}
\renewcommand{\arraystretch}{1.0}
\caption{\textbf{Results of detailed ablation study using the \emph{ground truth} 3D hand keypoints.} Experiments are performed on InterHand2.6M~\cite{moon2020interhand2} dataset.}
\vspace{-0.5\baselineskip}

\label{table:detailed_ablation_study}

\begin{tabularx}{\columnwidth}{>{\centering}m{4.8cm}|Y|Y}
\toprule
Method & IoU \footnotesize{(\%)} $\uparrow$ & CD \footnotesize{(mm)} $\downarrow$ \\
\midrule
$\mathcal{I} - \mathrm{ImageCond}$ & 74.7 & 2.62 \\
$\mathcal{I}$ - $\mathrm{QueryImageAtt}$ & 75.8 & 2.51 \\
$\mathcal{I}$ & 77.2 & 2.32 \\
\midrule
$\mathcal{I}$ + $\mathcal{R}$ - $\mathrm{InitOccCond}$ & 67.0 & 3.44 \\ 
$\mathcal{I}$ + $\mathcal{R}$ - $\mathrm{FeatureCloud}$ & 77.4 & 2.32 \\
$\mathcal{I}$ + $\mathcal{R}$ - $\mathrm{ContextLatent}$ & 77.6 & 2.31 \\
\midrule
\textbf{Im2Hands ($\mathcal{I}$ + $\mathcal{R}$)} & \textbf{77.8} & \textbf{2.30} \\ 
\bottomrule
\end{tabularx}
}
\end{table}

\noindent \textbf{Detailed Ablation Study With Predicted Keypoints.}
In Table~\ref{table:detailed_ablation_study_predicted_kpts}, we additionally show our results using the keypoints \emph{predicted} by DIGIT~\cite{fan2021learning} to examine the effectiveness of each of our components on more various settings. In the table, our full model is again shown to achieve the best performance. Considering the ablation study results using the ground truth (Table~\ref{table:detailed_ablation_study}) and predicted (Table~\ref{table:detailed_ablation_study_predicted_kpts}) keypoints together, one interesting observation is that $\mathcal{I}$ contributes more to the performance improvement when using the ground truth keypoints input, while $\mathcal{R}$ contributes more to it when using the \emph{predicted} keypoints input. It reveals that \emph{both} $\mathcal{I}$ and $\mathcal{R}$ are essential to enable robust two-hand shape reconstruction given input keypoints with various degrees of noise.


\begin{table}[!h]
\centering
{ \small
\setlength{\tabcolsep}{0.2em}
\renewcommand{\arraystretch}{1.0}
\caption{\textbf{Results of detailed ablation study using the 3D hand keypoints \emph{predicted} by DIGIT~\cite{fan2021learning}.} Experiments are performed on InterHand2.6M~\cite{moon2020interhand2} dataset.}
\label{table:detailed_ablation_study_predicted_kpts}

\begin{tabularx}{\columnwidth}{>{\centering}m{4.8cm}|Y|Y}
\toprule
Method & IoU \footnotesize{(\%)} $\uparrow$ & CD \footnotesize{(mm)} $\downarrow$ \\
\midrule
$\mathcal{K} + \mathcal{I} - \mathrm{ImageCond}$ & 53.0 & 5.63 \\
$\mathcal{K} + \mathcal{I}$ - $\mathrm{QueryImageAtt}$ & 53.9 & 5.47 \\
$\mathcal{K} + \mathcal{I}$ & 55.4 & 5.36 \\
\midrule
$\mathcal{K} + \mathcal{I}$ + $\mathcal{R}$ - $\mathrm{InitOccCond}$ & 55.1 & 5.18 \\ 
$\mathcal{K} + \mathcal{I}$ + $\mathcal{R}$ - $\mathrm{FeatureCloud}$ & 58.3 & 4.78 \\
$\mathcal{K} + \mathcal{I}$ + $\mathcal{R}$ - $\mathrm{ContextLatent}$ & 58.4 & 4.79 \\
\midrule
\textbf{Im2Hands ($\mathcal{K} + \mathcal{I}$ + $\mathcal{R}$)} & \textbf{59.4} & \textbf{4.75} \\ 
\bottomrule
\end{tabularx}
}
\end{table}

\noindent \textbf{Different Point Sampling Densities.} For our two-hand occupancy refinement, we represent each initial hand shape with 512 farthest points on the surface. In Table~\ref{table:rebuttal_sampling}, we also show the results with different point sampling densities. Our model performance (in $\mathrm{IoU}$) is not affected much by the point density, while the training time is reduced when the number of points is decreased -- showing the effectiveness of our method. Our model achieves state-of-the-art results even with the sparse 256 points.

\begin{table}[!h]
\centering
{ \small
\setlength{\tabcolsep}{0.2em}
\renewcommand{\arraystretch}{1.0}
\caption{\textbf{Training time (second per iteration) and IoU with varying point sampling density}. Time is obtained as an average of 1K measurements. For measuring IoU, we used the ground truth keypoint inputs.}
\label{table:rebuttal_sampling}
\vspace{-0.2\baselineskip}
\begin{tabularx}{\columnwidth}{>{\centering}m{3cm}|Y|Y}
\toprule
\# of sampled points & Training time (s) & IoU (\%) \\
\midrule
256 & \textbf{1.04} & 77.4 \\
\textbf{512 (Ours)} & 1.74 & \textbf{77.8} \\
1024 & 3.15 & 77.6 \\ 
\bottomrule
\end{tabularx}
}

\end{table}

\subsection{Implementation Details}
\label{subsec:implementation_details}

In this section, we report more details of our implementation that could not be included in the main paper due to the space limit. Note that more implementation details are also available through our code\footnote{\url{https://github.com/jyunlee/Im2Hands}}.

\subsubsection{Network Architecture}

\noindent \textbf{Initial Hand Occupancy Network ($\mathcal{I}$).}
For the query positional embedding module used to compute our query-image attention ($\mathrm{PosEnc}$ in Equation~\ref{eq:query-image-att}), we use a shared MLP composed of two fully-connected layers, each of them followed by ReLU activation and dropout with a rate of $0.01$. For the image encoder-decoder ($\mathrm{ImgEnc}$ in Equation~\ref{eq:query-image-att}), we use a ResNet-50~\cite{he2016deep} architecture as an encoder and a CNN composed of four deconvolutional layers as a decoder. For the multi-headed self-attention module ($\mathrm{MSA}$ in Equation~\ref{eq:query-image-att}), we extract features of $8 \times 8$ image patches using an encoder of Vision Transformer~\cite{dosovitskiy2020image} and apply self-attention with two attention heads. The resulting features extracted by query-image attention are concatenated with the features extracted by HALO~\cite{karunratanakul2021skeleton} encoder after the first layer in the part occupancy functions of HALO. For the architecture of HALO encoder and part occupancy functions, we follow the design of HALO. We thus refer the reader to \cite{karunratanakul2021skeleton} for more details.

\noindent \textbf{Two-Hand Occupancy Refinement Network ($\mathcal{R}$).}
For iso-surface point extraction, we evaluate the occupancy probabilities at uniformly sampled query points in 3D space and collect the query points that are estimated to be on the surface. We then apply farthest point sampling (FPS) to obtain 512 points to create each of the hand point clouds (i.e., $\mathcal{P}_l$ and $\mathcal{P}_r$). For feature cloud conversion, we use the same image encoder-decoder used in $\mathcal{I}$. For point cloud encoder ($\mathrm{PCEnc}$ in Equation~\ref{eq:pcd_encoding}), we use the same encoder architecture as in AIR-Net~\cite{giebenhain2021air} except for the input point dimension, which is increased due to our feature cloud conversion procedure. We use a shared $\mathrm{PCEnc}$ for both sides of hand feature clouds, but we distinguish each side by concatenating a binary label -- $[1, 0]$ for left hand and $[0, 1]$ for right hand -- to each of the point features. For our context encoder ($\mathrm{ContextEnc}$ in Equation~\ref{eq:context_encoding}), we concatenate the inputs ($z_l$, $z_r$, $z_I$) and apply an MLP composed of two fully-connected layers, each of them followed by ReLU activation. For our point cloud decoder that estimates the refined occupancy ($\mathrm{PCDec}$ in Equation~\ref{eq:pcd_decoding}), we concatenate the query coordinate $x$ with the initial occupancy probability at $x$ and feed the resulting query vector to the decoder of AIR-Net along with $\mathcal{A}_s$ and $z_c$. For more details on the architecture of $\mathrm{PCEnc}$ and $\mathrm{PCDec}$, please refer to \cite{giebenhain2021air}.

\noindent \textbf{Input Keypoint Refinement Network ($\mathcal{K}$).}
For $\mathrm{KptEnc}$, we use (1) an embedding layer to embed the index of each keypoint and (2) an MLP composed of two fully-connected layers to encode the coordinate of each keypoint. We then concatenate the index feature and the coordinate feature for each of the keypoints and set them as node features in a two-hand skeleton graph. We then feed the skeleton graph to a graph convolutional network ($\mathrm{GCN}$) composed of four layers with residual connections. The updated node features are directly used for multi-headed self-attention ($\mathrm{MSA}$) between the patch-wise image features, which are extracted by the same Vision Transformer~\cite{dosovitskiy2020image} encoder used in $\mathcal{I}$. The updated node features are then fed to an output keypoint coordinate regressor, which is an MLP composed of two fully-connected layers -- each of them followed by ReLU activation and dropout of a rate of $0.01$.


\subsubsection{Training Details}
\label{subsec:training_details}

For $\mathcal{I}$ and $\mathcal{R}$, we train each of the networks for 10 epochs with a batch size of 8. We use an Adam optimizer with an initial learning rate of $1e-4$, betas of $[0.9, 0.999]$, an epsilon of $1e-8$, and a weight decay parameter of $1e-5$. We additionally use a learning rate scheduler to decay the learning rate by $0.2$ every 5000 training steps. For the loss function to train $\mathcal{R}$, we use a weighted sum of the proposed loss terms (i.e., occupancy loss and penetration loss), with the weight values set as 1 and 0.001, respectively. Other training details (e.g., training query sampling) are the same as in the original HALO framework (please refer to \cite{karunratanakul2021skeleton} for more detail). For $\mathcal{K}$, we train the network for 30 epochs with a batch size of 32. We use an Adam optimizer with an initial learning rate of $1e-4$ with a scheduler to decay the learning rate by $0.3$ every 5000 training steps.

\subsection{Detailed Experimental Setups}
\label{subsec:detailed_experimental_setups}

\subsubsection{Metric Computation}

For Im2Hands and HALO~\cite{karunratanakul2020grasping}, we extract the reconstructed meshes by evaluating occupancy probabilities at uniformly sampled query points in 3D space and applying Marching Cubes~\cite{lorensen1987marching}. We then compute our metrics (i.e., mean Intersection over Union and Chamfer L1-Distance) after mid-joint alignment of each hand. Note that the existing works~\cite{li2022interacting, zhang2021interacting} use Mean Per Vertex Position Error (MPVPE) as an evaluation metric, which assumes one-to-one vertex correspondence between the ground truth and the predicted meshes. As our method does not assume such vertex correspondence, we use mean Intersection over Union and Chamfer L1-Distance as our evaluation metrics as in other implicit function-based reconstruction methods~\cite{karunratanakul2020grasping, deng2020nasa}.


\vspace{-0.5\baselineskip}
\subsubsection{Setups for Generalizability Test}

In this section, we report more details of our setups for the generalizability test (Section~\ref{subsec:generalizability_test} in the paper). For pre-processing the two-hand frames in RGB2Hands~\cite{wang2020rgb2hands} and EgoHands~\cite{bambach2015lending} datasets, we compute a coarse foreground mask obtained by thresholding the depth map provided by \cite{wang2020rgb2hands, bambach2015lending} to mask out the approximate background region. We then directly apply Im2Hands trained only on InterHand2.6M~\cite{moon2020interhand2} to evaluate its generalization ability to unseen hand shapes and appearances. Other experimental settings are the same as in our main experiments on InterHand2.6M dataset.

\end{document}